\documentclass[10pt,twocolumn,letterpaper]{article}

\usepackage[pagenumbers]{cvpr} % To force page numbers, e.g. for an arXiv version

\definecolor{cvprblue}{rgb}{0.21,0.49,0.74}
\usepackage[pagebackref,breaklinks,colorlinks,allcolors=cvprblue]{hyperref}
\usepackage[accsupp]{axessibility}
\usepackage{hyperref}
\usepackage{url}
\usepackage{graphicx}
\usepackage{booktabs}
\usepackage{xcolor}
\usepackage{multirow}
\usepackage{colortbl}
\usepackage{array}
\usepackage{wrapfig}
\usepackage{algorithm}
\usepackage{algpseudocode}
\usepackage{amsmath}
\usepackage{tabularx}
\usepackage{arydshln}
\usepackage{makecell}
\usepackage{amssymb}
\newcolumntype{Y}{>{\centering\arraybackslash}X}
\usepackage[table]{xcolor}
\newtheorem{lemma}{Lemma}
\definecolor{second_color}{HTML}{E4DEDB}
\definecolor{third_color}{HTML}{C9D2D7}
\definecolor{fourth_color}{HTML}{FFF1EA}

\title{PRISM: Video Dataset Condensation with \\ Progressive Refinement and Insertion for Sparse Motion}

\author{
Jaehyun Choi\textsuperscript{1,2},
Jiwan Hur\textsuperscript{2},
Gyojin Han\textsuperscript{2},
Jaemyung Yu\textsuperscript{3},
Junmo Kim\textsuperscript{2}\\
\textsuperscript{1}Korea AI Safety Institute, ETRI \quad
\textsuperscript{2}KAIST \quad
\textsuperscript{3}NAVER AI Lab\\
{\tt\small pre6ent@etri.re.kr, jaemyung.yu@navercorp.com,}\
{\tt\small \{jiwan.hur, hangj0820, junmo.kim\}@kaist.ac.kr}
}

\begin{document}

\maketitle
\begin{abstract}
Video dataset condensation aims to reduce the immense computational cost of video processing.
However, it faces a fundamental challenge regarding the inseparable interdependence between spatial appearance and temporal dynamics.
Prior work follows a static/dynamic disentanglement paradigm where videos are decomposed into static content and auxiliary motion signals.
This multi-stage approach often misrepresents the intrinsic coupling of real-world actions.
We introduce Progressive Refinement and Insertion for Sparse Motion (PRISM), a holistic approach that treats the video as a unified and fully coupled spatiotemporal structure from the outset.
To maximize representational efficiency, PRISM addresses the inherent temporal redundancy of video by avoiding fixed-frame optimization.
It begins with minimal temporal anchors and progressively inserts key-frames only where linear interpolation fails to capture non-linear dynamics.
These critical moments are identified through gradient misalignments.
Such an adaptive process ensures that representational capacity is allocated precisely where needed, minimizing storage requirements while preserving complex motion.
Extensive experiments demonstrate that PRISM achieves competitive performance across standard benchmarks while providing state-of-the-art storage efficiency through its sparse and holistically learned representation.
\end{abstract}     
\section{Introduction}
\label{sec:intro}

Modern machine learning research has progressed through the parallel development of novel algorithmic frameworks and the exponential growth of training data.
Within the field of computer vision, video stands as one of the most informative modalities because it captures the intricate evolution of spatial content over time.
Large-scale video datasets such as Kinetics-700~\citep{kinetics_2017},  HowTo100M~\citep{howto100m}, and YouTube-8M~\citep{youtubedata} have enabled remarkable advances in video understanding~\citep{kinetics_2017, intro2}, from action recognition, object tracking~\citep{intro3, intro4}, predicting future events~\citep{intro5}, to realistic video generation~\citep{intro6}.
However, the immense richness of these datasets comes at a steep price, introducing a critical bottleneck in storage, preprocessing, and training cost.
This escalating burden unintentionally limits broader participation in research in this field.
Although dataset condensation has proven effective in the image domain~\citep{dc_2020, dm_2023, idm, lossless_2023, mtt_2022, choi2025dam}, an analogous direction is now vital for the even greater computational demands of video.

Video dataset condensation is uniquely challenging due to the inseparable interdependence of content and motion.
Prior work~\citep{video_distillation_2024} typically disentangles these components into separate stages, but this decomposition often neglects the intrinsic spatiotemporal coupling of real-world sequences. For example, a single frame of hands meeting in a clapping action is visually indistinguishable from the moment hands begin to pull apart. This failure highlights a fundamental ambiguity where a static visual state can belong to multiple potential trajectories. Because content and motion are mutually constitutive, neither can be meaningfully defined in isolation. We therefore argue that video must be treated as a unified spatiotemporal entity from the beginning of the optimization. To maintain this holistic integrity while minimizing the memory footprint, we must avoid dense representations and instead focus on the most essential temporal anchors.

\begin{figure*}[t]
  \centering
   \includegraphics[width=0.8\textwidth]{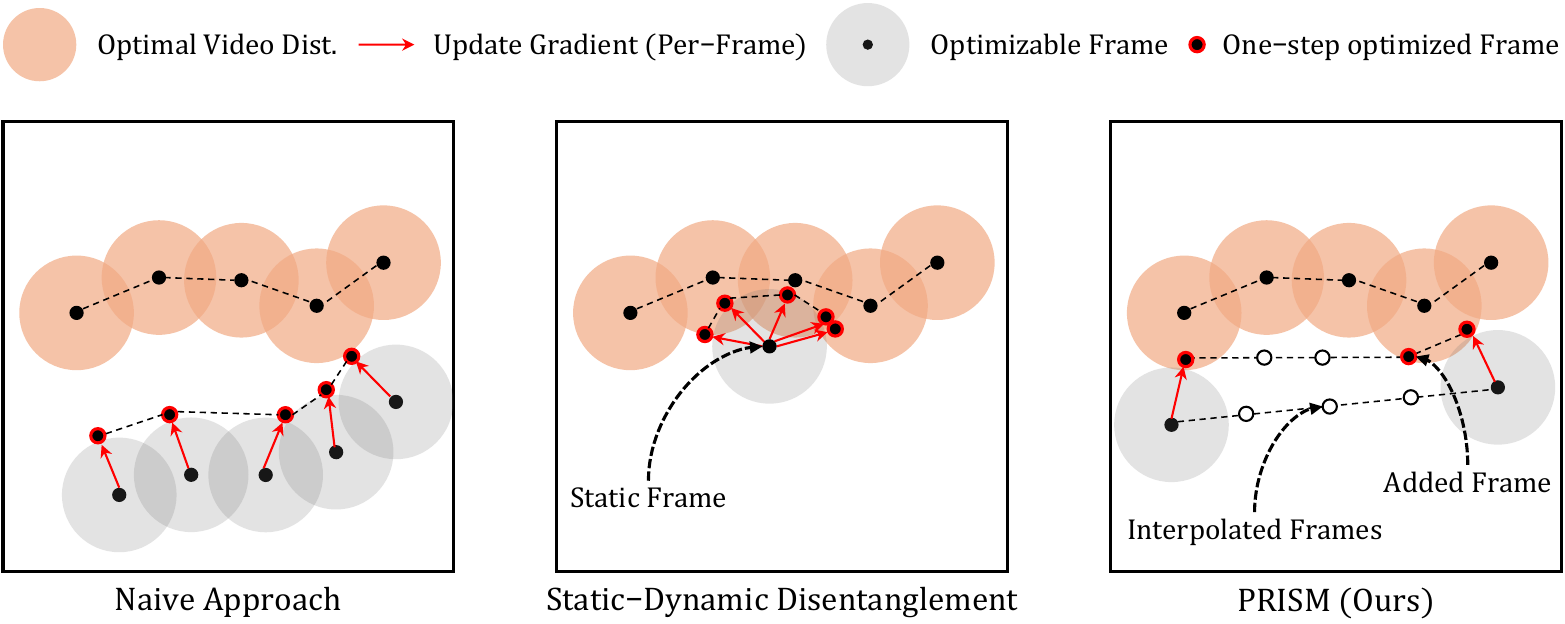}
   \vspace{-2mm}
   \caption{Visual representation of prior video dataset condensation methods and PRISM (Ours). In frame-wise matching, each frame gets updated individually, neglecting the relation between frames. Static and dynamic disentangling method~\cite{video_distillation_2024} learns the temporal dynamics; however, it is restricted by the frozen pre-trained static image. Unlike these methods, our method learns the motion dynamics without any constraints to a single frame through a holistic approach.}
   \vspace{-4mm}
   \label{fig:intro}
\end{figure*}

Based on this perspective, we propose Progressive Refinement and Insertion for Sparse Motion (PRISM).
Our method operationalizes the holistic approach by initializing each video with only two temporal anchors, specifically the start and the end frames. This minimal initialization ensures that the model begins with a global but sparse understanding of the action.
We assume that simple or low-velocity motion can be effectively approximated by linear interpolation between these anchors.
PRISM then focuses its representational capacity only on the frames that matter, which are the moments where this linear assumption fails to capture non-linear dynamics as shown in Figure ~\ref{fig:intro}.
These critical junctions are identified through gradient misalignments during the training process.
By adaptively inserting key-frames at these frames, PRISM preserves complex dynamics while achieving state-of-the-art storage efficiency.

Our main contributions are as follows:
\begin{itemize}
    \item We propose PRISM, a holistic paradigm for video dataset condensation that avoids artificial disentanglement and treats video as a unified spatiotemporal structure.
    \item We introduce a gradient-guided insertion mechanism that initializes with minimal anchors and adaptively identifies frames that are essential to understand the action and to capture non-linear dynamics.
    \item We demonstrate through extensive experiments that our method achieves highly competitive performance while establishing state-of-the-art storage efficiency across standard action recognition benchmarks.
\end{itemize}

\section{Related Works}
\label{sec:rel}

\subsection{Image Dataset Distillation} 

Image dataset condensation aims to synthesize a small set of representative samples that capture the essential characteristics of large-scale datasets.
The primary objective is to enable models trained on these condensed sets to achieve performance comparable to those trained on the full dataset, thereby significantly reducing computational and storage requirements.
This field has evolved into several distinct methodological branches.

\noindent\textbf{Gradient Matching} ensures that synthetic data produces gradient updates similar to those of the original dataset.
DC~\citep{dc_2020} pioneered this direction by formulating condensation as a bi-level optimization problem that matches single-step gradients.
DSA~\citep{dsa} enhanced this framework through differentiable Siamese augmentation to improve generalization across various transformations.
IDC~\citep{idc} further contributed efficient parameterization strategies by storing synthetic images at lower resolutions and upsampling during training, which effectively reduces storage requirements without compromising performance.

\noindent\textbf{Distribution Matching} aims to align feature distributions between synthetic and real data as an efficient alternative to gradient-based methods.
DM~\citep{dm_2023} introduced a framework that aligns distributions in embedding space to improve computational efficiency.
CAFE~\citep{cafe} ensures that statistical feature properties from synthetic and real samples remain consistent across network layers to provide more comprehensive feature alignment.
These methods typically offer better scaling properties when condensing large-scale datasets with numerous categories.

\noindent\textbf{Trajectory Matching} seeks to synchronize entire training trajectories rather than matching single-step gradients or feature distributions.
MTT~\citep{mtt_2022} developed techniques to create condensed datasets by mimicking the training trajectories of models trained on the original dataset.
DATM~\citep{lossless_2023} introduced difficulty-aligned trajectory matching to enable effective distillation without performance loss even as the synthetic dataset size changes.
Recently, the scope of this field has expanded to include multi-domain distillation to handle diverse data sources simultaneously~\citep{choi2025dam}.
While these techniques have achieved remarkable success in the static image modality, extending them to video sequences remains challenging due to the exponential increase in the search space and complex temporal dependencies.

\subsection{Video Dataset Condensation}
Despite the rapid progress in image condensation, research targeting the video modality remains sparse.
The only prior work in this area proposed a framework that disentangles video into static content and dynamic motion signals \citep{video_distillation_2024}.
This approach treats motion as an auxiliary component optimized separately from the spatial appearance through a two-stage process.
Furthermore, it relies on a fixed number of frames, which may incur unnecessary redundancy or fail to capture complex dynamics.
Although this strategy reduces initial computational complexity, it fundamentally misrepresents real-world actions where content and motion are mutually constitutive. 
PRISM differs from this paradigm by adopting a holistic learning framework that treats the video as a unified spatiotemporal structure from the outset.
Instead of utilizing dense or fixed frame sets, we introduce an adaptive representation that identifies the most informative temporal segments through gradient misalignment.
By beginning with minimal temporal anchors and progressively inserting frames only where linear assumptions fail, PRISM preserves action integrity while achieving state-of-the-art storage efficiency.
\section{Method}
\label{sec:method}

Let $\mathcal{D}=\bigcup_{c=0}^{C-1}\mathcal{D}_c$ denote the original large-scale dataset consisting of $C$ classes, where $\mathcal{D}_c =\{(V_c^i, y_c^i)\}_{i=1}^{N_c}$ represents the set of real videos for class $c$. 
Each video $V_c^i \in\mathbb{R}^{T\times H\times W\times 3}$ is a sequence of $T$ frames with height $H$, width $W$ and three color channels.
The goal of video dataset condensation is to synthesize a compact synthetic dataset:
\begin{equation}
    S = \bigcup_{c=0}^{C-1} S_c,\quad S_c = \{(S_c^j, y_c^j)\}_{j=1}^{M_c},\quad M_c \ll N_c,
\end{equation}
such that each condensed video $S_c^j$ captures the essential spatiotemporal dynamics specific to its respective class $c$, while drastically reducing memory and computation costs with minimal degradation in downstream task performance.

\subsection{Temporal Frame Interpolation}

Each synthetic sequence $S_c^j$ is parameterized by a sparse set of key-frames denoted as $\mathcal{K}_c^j=\{s_{c,k_1}, s_{c,k_2}, \dots, s_{c,k_n}\}$, where $1=k_1<k_2<\dots<k_n=T$ represent the discrete temporal indices.
At the beginning of the synthesis process, this set is initialized with $n=2$, using only the temporal boundaries of the video segment:
\begin{equation}
    \mathcal{K}_c^j = \{ s_{c,1}, s_{c,T} \}.
\end{equation}

This sparse representation assumes that simple or low-velocity motion can be effectively approximated through linear interpolation between key-frames~\citep{two_frame1, two_frame2}.
To reconstruct the full sequence $S_c^j$, all intermediate frames at indices $t\notin\{k_1,\dots,k_n\}$ are populated via linear interpolation. Given two adjacent key-frames $s_{c,k_i}$ and $s_{c, k_{i+1}}$, the interpolated frame at index t is computed as:
\begin{equation}
    s_{c,t} = \alpha_t s_{c,k_i} + (1 - \alpha_t) s_{c,k_{i+1}},
\end{equation}
where $ \alpha_t = \frac{k_{i+1} - t}{k_{i+1} - k_i}$ and $ k_i < t < k_{i+1}$.
Throughout the optimization process, only the frames designated within the key-frame set $\mathcal{K}_c^j$ serve as trainable parameters, while the remaining frames in $S_c^j$ are updated indirectly through the interpolation mechanism.
This parameterization enables a holistic spatiotemporal coupling by optimizing the entire sequence within a unified framework.

\begin{figure*}[t]
  \centering
   \includegraphics[width=\textwidth]{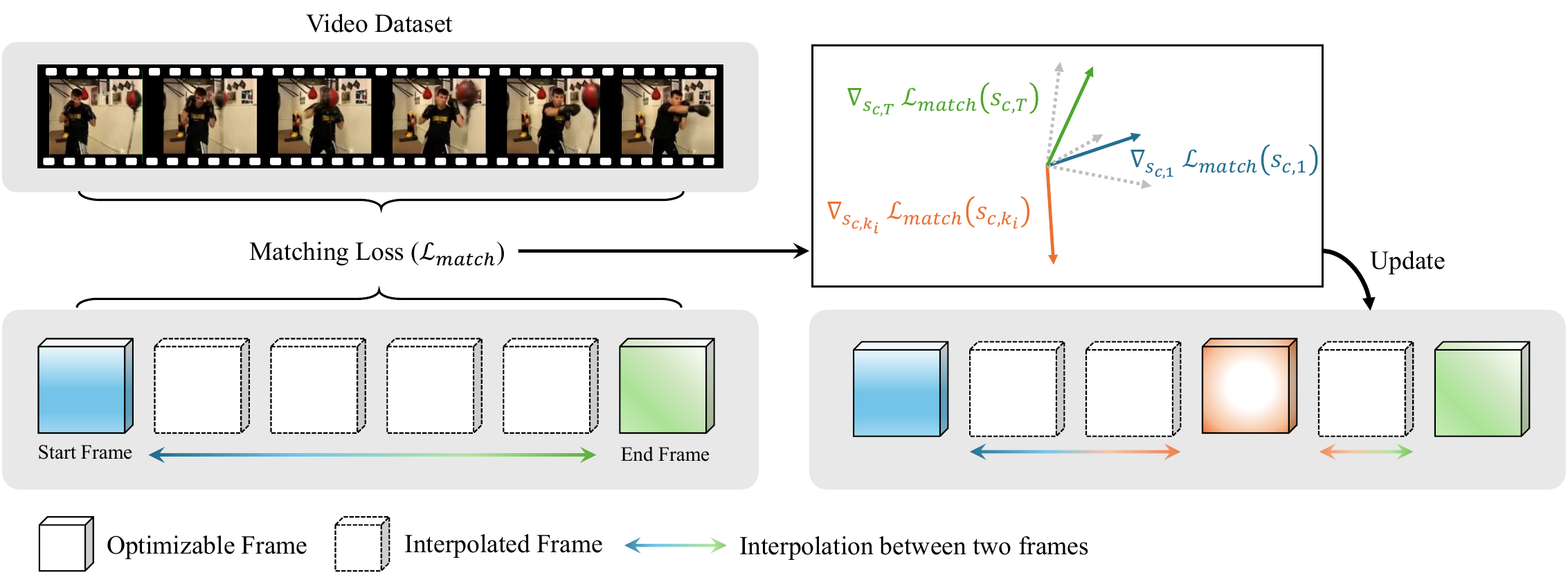}
   \vspace{-2mm}
   \caption{\textbf{Overview of PRISM.} We first initialize the key-frame set as the start and the end frame. 
   As training proceeds, we calculate the cosine similarity of the gradients for each temporally interpolated frame against its two adjacent key-frames.
   Then, the frames that have a negative correlation with its two adjacent key-frames are added to the key-frame set and used in the subsequent training.
   }
   \vspace{-4mm}
   \label{fig:main}
\end{figure*}

\subsection{Gradient-Guided Frame Insertion}

The initial set $\mathcal{K}_c^j$ is progressively expanded by identifying and inserting frames at indices where the linear approximation fails to capture complex spatiotemporal dynamics.
The necessity of this insertion is grounded in the theoretical observation that certain non-linear transitions create a state where updates to the existing key-frames cannot reduce the loss at intermediate points.
This phenomenon is formalized in the following lemma.

\begin{lemma}
\label{lem:loss_conflict}
\textnormal{\textbf{(Loss-Descent Conflict under Gradient Misalignment)}}
\quad Let $s_t = \alpha s_{k_i} + (1 - \alpha) s_{k_{i+1}}$ be a linear interpolation between two key-frames $s_{k_i}$ and $s_{k_{i+1}}$ for $0 < \alpha < 1$.
Let the task-loss gradients be
\[
g_t = \nabla_{s_t} \mathcal{L}(s_t),\; 
g_i = \nabla_{s_{k_i}} \mathcal{L}(s_{k_i}),\; 
g_{i+1} = \nabla_{s_{k_{i+1}}} \mathcal{L}(s_{k_{i+1}}).
\]
Suppose 
$
\langle g_t, g_i \rangle < 0 \quad \text{and} \quad \langle g_t, g_{i+1} \rangle < 0.
$
Then, for any convex combination
\[
v = \lambda (-g_i) + (1 - \lambda)(-g_{i+1}), \quad \lambda \in [0, 1],
\]
it holds that
\[
\langle g_t, v \rangle > 0.
\]
Consequently, no first-order update to only the endpoints decreases $\mathcal{L}(s_t)$.
\end{lemma}
\emph{Proof.} See Appendix. \hfill $\square$

Lemma 1 indicates that when the gradient of an intermediate frame is directionally misaligned with those of its anchors, the current interpolation baseline is insufficient for further loss minimization of the sequence.
To resolve this, such frames must be promoted to the key-frame set to be optimized directly. In practice, this inadequacy is detected by monitoring the directional alignment of the gradients during training. For all candidate frames $s_{c,t}$ between adjacent key-frames $s_{c,k_i}$ and $s_{c,k_{i+1}}$, the alignment is quantified via cosine similarity:
\begin{equation}
    \cos_i^t = \cos\left( \nabla \mathcal{L}(s_{c,t}), \nabla \mathcal{L}(s_{c,k_i}) \right),
\end{equation}
\begin{equation}
    \cos_{i+1}^t = \cos\left( \nabla \mathcal{L}(s_{c,t}), \nabla \mathcal{L}(s_{c,k_{i+1}}) \right).
\end{equation}
A frame $s_{c,t}$ is identified as a critical temporal junction where the linear assumption is insufficient if both similarities fall below a predefined threshold $\epsilon$:
\begin{equation}
\cos_i^t < \epsilon \quad \text{and} \quad \cos_{i+1}^t < \epsilon.
\end{equation}
Upon satisfaction of this criterion, the candidate frame is inserted into the set such that 
\begin{equation}
    \mathcal{K}_c^j \leftarrow \mathcal{K}_c^j \cup \{ s_{c,t} \}.
\end{equation}

This gradient-based criterion captures non-linear transitions in appearance or motion and enables the model to refine its support set by inserting only those frames that contribute meaningful learning signals.
The insertion process is repeated iteratively throughout training except for the warm-up and cool-down phases, resulting in a temporally adaptive sequence that emphasizes semantically rich regions.
To ensure stable selection and resolve dependencies between adjacent candidate frames, an iteration-wise update strategy is employed.
Within a single training iteration, the current key-frame set is kept fixed while we evaluate the gradient conflict criterion for all candidate frames.
Candidates that satisfy the negative cosine similarity condition are inserted into the key-frame set simultaneously at the end of the iteration. In the subsequent iteration, the temporal interpolation and gradients are recomputed based on this newly updated key-frame set.
This ensures that the insertion of a frame properly influences the evaluation of surrounding candidates in the next step, maintaining optimization stability.

\subsection{Warm-Up and Cool-Down Phase}

To ensure stable frame insertion dynamics during training, a warm-up and cool-down phases are introduced.

\noindent\textbf{Warm-up Phase.}
During the early stage of training, gradient-guided frame insertion is disabled, and optimization is restricted to the initial key-frames in $\mathcal{K}_c^j = \{s_{c,1}, s_{c,T}\}$.
This phase allows the temporal endpoints to stabilize before serving as reference anchors for subsequent insertion decisions.
Without this stabilization phase, premature insertion driven by noisy early gradients may lead to the selection of redundant or suboptimal key-frames.

\noindent\textbf{Cool-Down Phase.}
In the final stage of training, frame insertion is again disabled.
This phase accounts for the fact that frames inserted near convergence receive insufficient updates, potentially remaining under-optimized and degrading downstream performance.
By freezing the key-frame set toward the end of training, adequate supervision is ensured for all selected frames.
In practice, both the warm-up and cool-down phases are allocated for 20\% of the total iterations each, leaving the middle 60\% of the iterations for the progressive refinement and insertion process.

\subsection{Optimization Objective}
\label{sec:3.4}
Let $f_\theta$ denote a 3D convolutional feature extractor.
For a given class $c$, the discrepancy between the real video batch $\mathcal{B}_c^{real}$ and the synthetic video batch $\mathcal{B}_c^{syn}$ is minimized through the following objective function:
\begin{equation}
    \min_{\mathcal{\mathcal{K}}} \sum_{c=0}^{C-1} \left\| \frac{1}{|\mathcal{B}_c^{real}|} \sum_{x \in \mathcal{B}_c^{real}} f_\theta(x) - \frac{1}{|\mathcal{B}_c^{syn}|} \sum_{s \in \mathcal{B}_c^{syn}} f_\theta(s) \right\|^2,
\end{equation}
where $\mathcal{K} = \bigcup_{c,j} \mathcal{K}_c^j$ represents the global set of all trainable key-frames.

\begin{table*}[ht!]
    \caption{Experiment results on two video benchmarks and prior methods categorized into coreset methods, static/dynamic disentangled methods, and holistic methods. $\textsuperscript{\dag}$ represents the author provided results. Higher values are better.
    \textbf{Bold} and \underline{underline} denote the best and second-best scores for each setting, respectively.}
    \vspace{-2mm}
    \centering
    {\small
    \begin{tabularx}{\textwidth}{lYYYYYY}
    \toprule
    \multirow{2}{*}{\textbf{Method}} & \multicolumn{3}{c}{\textbf{MiniUCF}} & \multicolumn{3}{c}{\textbf{HMDB51}} \\
    \cmidrule(lr){2-4} \cmidrule(lr){5-7}
     & VPC 10 & VPC 5 & VPC 1 & VPC 10 & VPC 5 & VPC 1 \\
    \midrule
    \midrule
            \multicolumn{7}{l}{\cellcolor{fourth_color}\textbf{Coreset Methods}} \\
            \midrule
            
            Random & $27.8_{\pm1.1}$ & $19.6_{\pm0.4}$ & $10.9_{\pm0.7}$ & $9.8_{\pm0.4}$ & $6.8_{\pm0.7}$ & $3.3_{\pm0.1}$ \\
            Herding & $\textbf{33.7}_{\pm0.3}$ & $26.3_{\pm1.0}$ & $13.2_{\pm1.3}$ & $10.8_{\pm0.6}$ & $9.0_{\pm0.6}$ & $3.0_{\pm0.1}$ \\
            K-Center & $29.1_{\pm0.6}$ & $23.2_{\pm0.7}$ & $13.9_{\pm1.6}$ & $8.0_{\pm0.1}$ &  $5.2_{\pm0.4}$ & $2.4_{\pm0.4}$ \\
            
            \midrule
            \multicolumn{7}{l}{\cellcolor{second_color}\textbf{Static / Dynamic Disentangled Methods}} \\
            \midrule
            
            DM & $30.0_{\pm0.6}$ & $25.7_{\pm0.2}$ & $15.3_{\pm1.1}$ & $\underline{12.1}_{\pm0.4}$ & $8.0_{\pm0.2}$ & $\underline{6.1}_{\pm0.2}$ \\
            Wang et al.$\textsuperscript{\dag}$ & - & $\underline{27.2}_{\pm0.4}$ & $\underline{17.5}_{\pm0.1}$ & - & $\underline{8.2}_{\pm0.1}$ & $6.0_{\pm0.4}$ \\
            
            \midrule
            \multicolumn{7}{l}{\cellcolor{third_color}\textbf{Holistic Method}} \\
            \midrule
            
            \textbf{PRISM} & $\underline{31.0}_{\pm0.1}$ & $\textbf{28.0}_{\pm0.1}$ & $\textbf{17.9}_{\pm0.3}$ & $\textbf{12.8}_{\pm0.2}$ & $\textbf{10.5}_{\pm0.4}$ & $\textbf{7.5}_{\pm0.3}$ \\
            
            \midrule
            \midrule
            
            Whole Dataset & \multicolumn{3}{c}{$57.8_{\pm1.1}$} & \multicolumn{3}{c}{$25.4_{\pm0.2}$} \\
            
            \bottomrule
        \end{tabularx}
    }
    \vspace{-2mm}
    \label{tab:main_performance}
\end{table*}
\begin{table*}[ht!]
    \caption{Storage requirements in MB on two video benchmarks and prior methods categorized into coreset methods, static/dynamic disentangled methods, and holistic methods. $\textsuperscript{\dag}$ represents the author provided results. Lower values are better.
    The value in parentheses indicates the size of the synthetic dataset as a percentage of the full dataset size.
    \textbf{Bold} and \underline{underline} denote the best and second-best scores for each setting, respectively.}
    \vspace{-2mm}
    \centering
    \small
    {\begin{tabularx}{\textwidth}{lYYYYYY}
    \toprule
    \multirow{2}{*}{\textbf{Method}} & \multicolumn{3}{c}{\textbf{MiniUCF}} & \multicolumn{3}{c}{\textbf{HMDB51}} \\
    \cmidrule(lr){2-4} \cmidrule(lr){5-7}
     & VPC 10 & VPC 5 & VPC 1 & VPC 10 & VPC 5 & VPC 1 \\
    \midrule
    \midrule
            \multicolumn{7}{l}{\cellcolor{fourth_color}\textbf{Coreset Methods}} \\
            \midrule
            
            Random & $1150_{(11.7\%)}$ & $586_{(6.0\%)}$ & $115_{(1.2\%)}$ & $1150_{(23.3\%)}$ & $586_{(11.9\%)}$ & $115_{(2.3\%)}$ \\
            Herding & $1150_{(11.7\%)}$ & $586_{(6.0\%)}$ & $115_{(1.2\%)}$ & $1150_{(23.3\%)}$ & $586_{(11.9\%)}$ & $115_{(2.3\%)}$ \\
            K-Center & $1150_{(11.7\%)}$ & $586_{(6.0\%)}$ & $115_{(1.2\%)}$ & $1150_{(23.3\%)}$ & $586_{(11.9\%)}$ & $115_{(2.3\%)}$ \\
            
            \midrule
            \multicolumn{7}{l}{\cellcolor{second_color}\textbf{Static / Dynamic Disentangled Methods}} \\
            \midrule
            
            DM& $1150_{(11.7\%)}$ & $586_{(6.0\%)}$ & $115_{(1.2\%)}$ & $1150_{(23.3\%)}$ & $586_{(11.9\%)}$ & $115_{(2.3\%)}$\\
            + Wang et al.$\textsuperscript{\dag}$ & - & $455_{(4.6\%)}$ & $94_{(1.0\%)}$ & - & $455_{(9.2\%)}$ & $94_{(1.9\%)}$ \\
            
            \midrule
            \multicolumn{7}{l}{\cellcolor{third_color}\textbf{Holistic Methods}} \\
            \midrule
            
            \textbf{PRISM} & $\textbf{324}_{(3.3\%)}$ & $\textbf{133}_{(1.4\%)}$ & $\textbf{20}_{(0.2\%)}$ & $\textbf{287}_{(5.8\%)}$ & $\textbf{137}_{(2.8\%)}$ & $\textbf{22}_{(0.4\%)}$ \\
            
            \midrule
            \midrule
            
            Whole Dataset & \multicolumn{3}{c}{9.81GB} & \multicolumn{3}{c}{4.93GB} \\
            
            \bottomrule
        \end{tabularx}
    }
    \vspace{-4mm}
    \label{tab:main_storage}
\end{table*}

While the loss is computed over the full reconstructed sequences $s \in \mathcal{B}_c^{syn}$, which consist of both key-frames and interpolated frames, the gradients are backpropagated exclusively to the sparse key-frame set $\mathcal{K}_c^j$.
This targeted update mechanism ensures that the optimization remains focused on the representative anchors while the holistic spatiotemporal structure is preserved through the interpolation framework.
The overall framework of the PRISM is illustrated in Figure~\ref{fig:main}.

\section{Experiment}
\label{sec:exp}

\subsection{Dataset}
We conduct experiments on 4 datasets: UCF101~\citep{ucf101_2012} and HMDB51~\citep{hmdb_2011} for small scale datasets, and  Kinetics~\citep{kinetics_2017} and Something-Something V2~\citep{ssv2_2017} for large scale datasets.
UCF101 consists of 13,320 video clips of 101 action categories.
Following the prior work~\citep{video_distillation_2024}, we leverage the miniaturized version of UCF101, hereinafter miniUCF, which includes the 50 most common action categories from the UCF101 dataset.
HMDB51 consists of 6,849 video clips of 51 action categories.
Kinetics-400 has videos of 400 human action classes and Something-Something V2 has 174 motion-centered classes.

For miniUCF and HMDB51, we sample 16 frames per video with a sampling interval of 4 and resize frames to $112\times112$.
For Kinetics-400 and Something-Something V2, we sample 8 frames per video and resize to $64\times64$.
Consistent with prior work~\citep{video_distillation_2024}, we only apply horizontal flipping with 50\% probability as the sole data augmentation strategy.

\subsection{Experimental Setting}

\noindent\textbf{Backbone and Metric.}
Following the established protocol in the pioneering work of video dataset condensation, we employ miniC3D, comprising 4 Conv3D layers, as our default backbone for all training stages.
We report the mean of three independent evaluations to ensure statistical significance.
For miniUCF and HMDB51, we measure top-1 accuracy, while for Kinetics-400 and Something-Something V2, we utilize top-5 accuracy.

\noindent\textbf{Initialization and Training.}
One of the distinct differences with prior video dataset condensation is our choice of initialization.
While methods like DM~\citep{dm_2023} typically initialize synthetic samples from random real frames, we initialize PRISM from Gaussian noise.
This choice reflects our holistic philosophy by allowing the model to synthesize the entire spatiotemporal structure from scratch, guided by the dataset's distribution rather than starting from a pre-existing spatial state.

\noindent\textbf{Baselines.}
We benchmark PRISM against three coreset selection methods (random selection, Herding~\citep{herding_2009}, and K-Center~\citep{kcenter-2017}), an image dataset condensation method (DM~\citep{dm_2023}), and a video dataset condensation method (Wang et al.~\citep{video_distillation_2024}) the first and the only video dataset condensation method.
We evaluate performance under varying Videos Per Class (VPC).
Note that the VPC follows the notation of Images Per Class (IPC) in image dataset condensation and that PRISM, in most cases, will have fewer frames than 16 frames, as we are only adding frames when required.
During inference, we leverage the index position for each saved vector, which is saved with the frames, with negligible memory consumption.
Our experiments employ the SGD optimizer with a momentum of 0.95 for all methods.
The $\epsilon$ is set to 0.
We found this value to be a robust default, as our gradient-guided criterion is primarily concerned with directional misalignment rather than its precise magnitude.
A detailed sensitivity analysis on the choice of $\epsilon$ is provided in the supplementary material along with other detailed analysis.

\begin{table}[t!]

\caption{Experiment results on two large-scale video benchmarks. $\textsuperscript{\dag}$ represents the author provided results.
\textbf{Bold} and \underline{underline} denote the best and second-best scores for each setting, respectively.}
\vspace{-2mm}
\centering
\small
\begin{tabularx}{\columnwidth}{lYYYY}
\toprule
\multirow{2}{*}{\textbf{Method}} & \multicolumn{2}{c}{\textbf{Kinetics-400}} & \multicolumn{2}{c}{\textbf{SSv2}} \\
\cmidrule(lr){2-3} \cmidrule(lr){4-5}
 & VPC 5 & VPC 1 & VPC 5 & VPC 1 \\
\midrule
\midrule
        \multicolumn{5}{l}{\cellcolor{fourth_color}\textbf{Coreset Methods}} \\
        \midrule
        
        Random & $5.5_{\pm0.2}$ & $3.0_{\pm0.2}$ & $3.6_{\pm0.1}$ & $3.1_{\pm0.1}$ \\
        Herding & $6.3_{\pm0.2}$ & $3.3_{\pm0.1}$ & $3.6_{\pm0.1}$ & $2.8_{\pm0.1}$ \\
        K-Center & $6.2_{\pm0.2}$ & $3.1_{\pm0.1}$ & $\textbf{4.5}_{\pm0.1}$ & $2.6_{\pm0.2}$ \\
        
        \midrule
        \multicolumn{5}{l}{\cellcolor{second_color}\textbf{Static / Dynamic Disentangled Methods}} \\
        \midrule
        
        DM & $\textbf{9.1}_{\pm0.9}$ & $\underline{6.3}_{\pm0.0}$ & $\underline{4.1}_{\pm0.0}$ & $3.6_{\pm0.0}$ \\
        + Wang et al.$\textsuperscript{\dag}$ & $7.0_{\pm0.1}$ & $\underline{6.3}_{\pm0.2}$ & $3.8_{\pm0.1}$ & $\textbf{4.0}_{\pm0.1}$ \\
        
        \midrule
        \multicolumn{5}{l}{\cellcolor{third_color}\textbf{Holistic Method}} \\
        \midrule
        
        \textbf{PRISM} & $\underline{8.1}_{\pm0.1}$ & $\textbf{7.1}_{\pm0.1}$ & $\underline{4.1}_{\pm0.1}$ & $\underline{3.9}_{\pm0.2}$ \\
        
        \midrule
        \midrule
        
        Whole Dataset & \multicolumn{2}{c}{$34.6_{\pm0.5}$} & \multicolumn{2}{c}{$29.0_{\pm0.6}$} \\
        
        \bottomrule
\end{tabularx}
\vspace{-4mm}
\label{tab:large-scale}
\end{table}

\subsection{Results}
\noindent\textbf{Performance Analysis.}
Table~\ref{tab:main_performance} presents the experiment results categorized as coreset methods, static and dynamic disentangled methods, and the holistic method.
We categorize DM~\citep{dm_2023} as a disentangled approach because it initializes synthetic frames from real dataset samples.
Overall, the results demonstrate that PRISM achieves highly competitive performance across various settings, often outperforming prior methods while establishing state-of-the-art storage efficiency.
On the large-scale Kinetics-400 dataset, the performance increments are naturally smaller since only 8 frames are utilized for training.
The performance increments of PRISM are smaller in the large dataset, where only 8 frames are used for training, as shown in Table~\ref{tab:large-scale}.
We observe that in the extremely low-data regime of Kinetics-400 at VPC 5 ($T=8$, $64\times64$), the image-based DM baseline outperforms PRISM ($9.1\pm0.9$ vs. $8.1\pm0.1$).
This specific performance gap may be attributed to the inherent stability of image-based initialization compared to our method's dynamic synthesis from noise.
However, even in this challenging environment, PRISM still outperforms the only other video-specific method (Wang et al., $7.0\pm0.1$) and successfully regains the top performance at VPC 1 ($7.1\pm0.1$ vs. DM’s $6.3\pm0.0$).
This shift demonstrates PRISM’s strong efficacy and relevance in high-compression scenarios.

\noindent\textbf{Storage and Representation Efficiency.}
As the storage of condensed data is a critical factor in dataset condensation, we report the storage requirements in Table~\ref{tab:main_storage}.
We follow the same calculation procedure as prior work~\citep{video_distillation_2024}, treating each synthetic sample as a \texttt{float32} tensor to determine the total memory usage.
For PRISM, the reported storage corresponds to the total number of frames retained after the condensation process completes for each VPC setting, while we ignore the negligible overhead from storing frame indices.
Unlike previous methods that begin with a fixed number of frames (e.g., 16), PRISM starts with only 2 key-frames per video and progressively inserts additional frames only when the cosine similarity between gradients is negative.
This selective strategy results in a significantly lower storage while achieving superior performance.
Specifically, on the miniUCF dataset at 1 VPC, PRISM achieves higher accuracy ($17.9\%$) than the prior distillation baseline ($17.5\%$) while requiring only 20 MB of storage, a nearly five-fold reduction compared to the baseline's 94 MB.
Moreover, since PRISM adds frames adaptively rather than proportionally to the number of VPCs, its storage does not grow linearly with VPC.
This behavior is clearly visible in the miniUCF results, where storage grows much more slowly than would be expected under proportional expansion.
Such efficiency makes PRISM especially advantageous when users wish to scale up performance under higher VPC budgets without incurring prohibitive storage costs.
\section{Ablation}
\label{sec:ablation}
\begin{table}[t!]
\caption{Cross-architecture results on MiniUCF with 1 VPC. \textsuperscript{\dag} indicates author-provided results. \textbf{Bold} denotes best scores.}
\vspace{-2mm}
\centering
\small
\begin{tabularx}{\columnwidth}{lYYY}
        \toprule
        \multirow{1}{*}{Method} & \multicolumn{3}{c}{Evaluation Model} \\
        \cmidrule(lr){2-4}
        & ConvNet3D & CNN+GRU & CNN+LSTM \\
        \midrule
        DM & $15.3_{\pm1.1}$ & $9.9_{\pm0.7}$  & $9.2_{\pm0.3}$ \\
        Wang et al.\textsuperscript{\dag} & $17.5_{\pm0.1}$ & $12.0_{\pm0.7}$ & $10.3_{\pm0.2}$ \\
        PRISM (Ours) & $\textbf{17.9}_{\pm0.3}$ & $\textbf{18.9}_{\pm0.8}$ & $\textbf{18.2}_{\pm1.3}$ \\
        \bottomrule
\end{tabularx}
\vspace{-4mm}
\label{tab:cross_arch}
\end{table}

\noindent\textbf{Cross-Architecture Generalization.}
A critical requirement for dataset condensation is the ability of the synthesized samples to perform well on architectures beyond the training backbone.
We validate the robustness of PRISM by evaluating our 1 VPC miniUCF condensed set on three different models: ConvNet3D, CNN+GRU, and CNN+LSTM.
As shown in Table \ref{tab:cross_arch}, PRISM not only achieves highly competitive performance compared to prior methods but also maintains significantly more robust performance across these diverse architectures compared to prior methods.
We presume that this robust generalization is a direct benefit of our gradient-guided frame insertion.
By restricting the  optimization exclusively to essential key-frames identified through gradient misalignment, the direct influence of the training backbone's specific inductive bias is reduced.
Unlike dense methods that optimize every frame, potentially overfitting to the specific architecture, PRISM focuses on learning a sparse set of fundamental spatiotemporal anchors.
This approach ensures that the intrinsic structure of the action is preserved through holistic coupling rather than auxiliary signals, making the synthesized data more adaptable to different modeling paradigms.

\begin{figure}[t!]
  \centering
   \includegraphics[width=1.0\columnwidth]{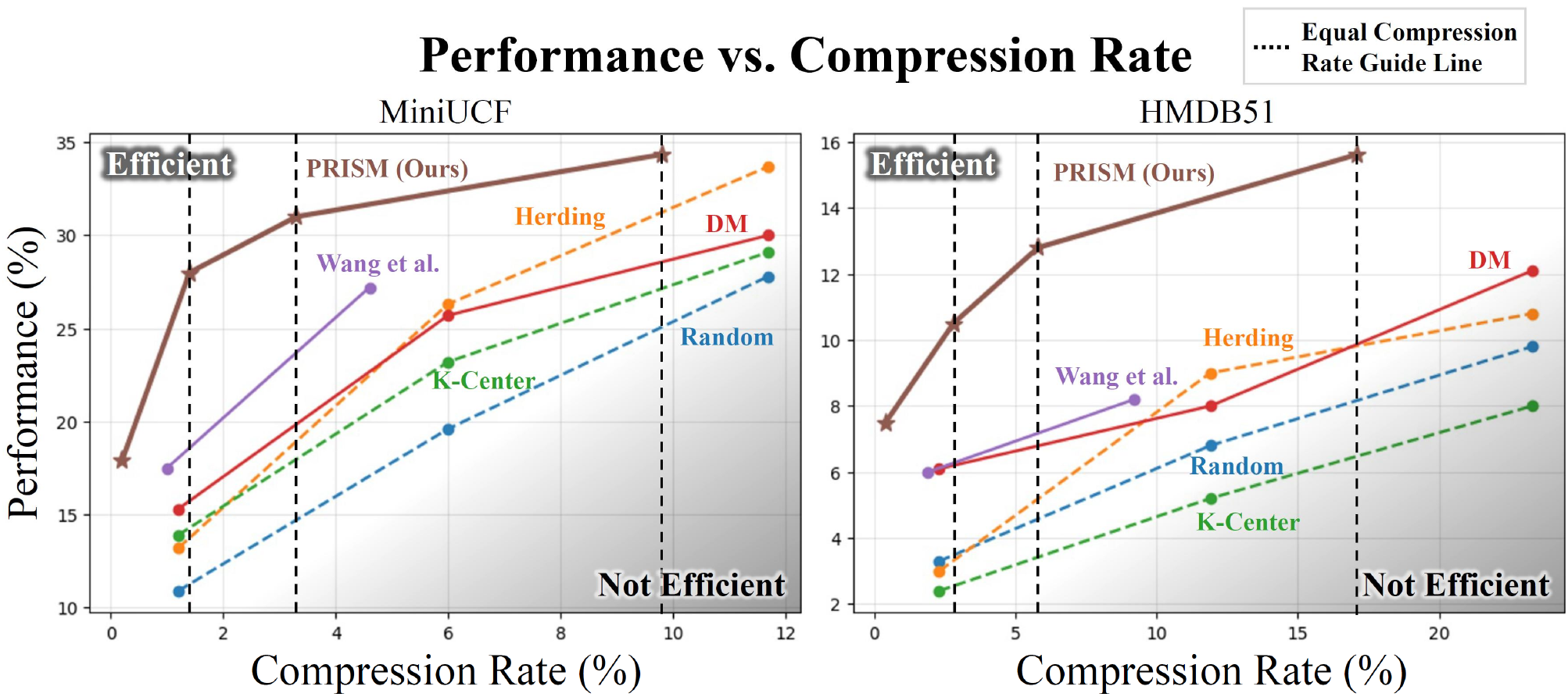}
   \caption{Accuracy-compression results.}
   \label{fig:rebuttal_storage_budget_comparison}
\vspace{-2mm}
\end{figure}

\noindent\textbf{Scaling under Equivalent Storage.}
Video dataset condensation performance is traditionally reported against a fixed number of Videos Per Class (VPC)  to maintain consistency with prior protocols.
However, this metric does not account for the actual storage of the condensed set.
We therefore shift our standard of comparison from VPC to Compression Rate, defined as the synthetic dataset size relative to the full dataset represented by the percentage values in Table~\ref{tab:main_storage}.
Coreset methods may appear competitive in raw accuracy under identical VPC settings, but they incur a significantly higher storage cost due to their reliance on dense, 16-frame segments as shown in Figure~\ref{fig:rebuttal_storage_budget_comparison}.
We demonstrate that PRISM remains superior when evaluated by actual storage requirements by relaxing our compression constraint and increasing the budget to 30 VPC.
PRISM achieves accuracies of 34.4\% on miniUCF and 15.6\% on HMDB51 in this setting.
Notably, even at 30 VPC, our overall compression rate on miniUCF and HMDB51 remains lower than that of baseline coreset methods at only 10 VPC.
This comparison proves that shifting the evaluation metric to storage efficiency allows PRISM to effectively scale its representational capacity and achieve much higher performance than fixed-budget alternatives.

\begin{table}[t!]
\centering
\caption{Results by varying initial key-frames of PRISM. Bold denotes best scores per dataset.}
\vspace{-2mm}
\small
\begin{tabularx}{\columnwidth}{lYYYYY}
        \toprule
        \multirow{2}{*}{Dataset} & \multicolumn{5}{c}{Number of Initial Key-Frames} \\
        \cmidrule{2-6}
        & 2 & 3 & 4 & 6 & 8 \\
        \midrule
        MiniUCF & $\textbf{17.9}_{\pm0.3}$ & $17.3_{\pm0.8}$ & $17.0_{\pm0.1}$ & $15.5_{\pm0.2}$ & $15.3_{\pm0.2}$\\
        HMDB51 & $\textbf{7.5}_{\pm0.3}$ & $6.8_{\pm0.1}$ & $6.6_{\pm0.1}$ & $6.1_{\pm0.3}$ & $5.6_{\pm0.2}$ \\
        \bottomrule
\end{tabularx}
\vspace{-4mm}
\label{tab:start_num_frame}
\end{table}
\begin{table*}[t!]

\caption{Ablation study results for (A) with and without insertion, (B) frame selection strategy, (C) similarity metric, and (D) training phase (warm-up and cool-down).}
\vspace{-2mm}
\centering
\begin{minipage}{0.49\textwidth}
\small
\centering
\begin{tabularx}{\textwidth}{YYY}
\toprule
Dataset & w/ Insertion & w/o Insertion \\
\midrule
HMDB51  & $7.5_{\pm0.3}$ & $6.1_{\pm0.3}$ \\
[0.1ex] \cdashline{1-3}[3pt/3pt] \\ [-1.8ex]
MiniUCF  & $17.9_{\pm0.3}$ & $15.8_{\pm1.2}$ \\
\bottomrule
\end{tabularx}
\\[1.0ex](A)\\[1.0ex]
\end{minipage}
\hfill
\vspace{0.5em}
\begin{minipage}{0.49\textwidth}
\small
\centering
\begin{tabularx}{\textwidth}{YYY}
\toprule
Dataset & Negative Grad. & Random Pos. \\
\midrule
HMDB51 & $7.5_{\pm0.3}$ & $6.8_{\pm0.2}$ \\
[0.1ex] \cdashline{1-3}[3pt/3pt] \\ [-1.8ex]
MiniUCF & $17.9_{\pm0.3}$ & $16.8_{\pm0.4}$ \\
\bottomrule
\end{tabularx}
\\[1.0ex](B)\\[1.0ex]
\end{minipage}
\begin{minipage}{0.49\textwidth}
\small
\centering
\begin{tabularx}{\textwidth}{YYY}
\toprule
Dataset & Cosine Sim. & L2 Distance \\
\midrule
HMDB51 & $7.5_{\pm0.3}$ & $6.0_{\pm0.6}$ \\
[0.1ex] \cdashline{1-3}[3pt/3pt] \\ [-1.8ex]
MiniUCF & $17.9_{\pm0.3}$ & $15.7_{\pm0.7}$ \\
\bottomrule
\end{tabularx}
\\[1.0ex](C)
\end{minipage}
\hfill
\begin{minipage}{0.49\textwidth}
\small
\centering
\begin{tabularx}{\textwidth}{YYY}
\toprule
Dataset & w/o Warm-Up & w/o Cool-Down \\
\midrule
HMDB51 & $6.8_{\pm1.2}$ & $6.3_{\pm0.3}$ \\
[0.1ex] \cdashline{1-3}[3pt/3pt] \\ [-1.8ex]
MiniUCF & $16.1_{\pm0.8}$ & $16.9_{\pm1.3}$ \\
\bottomrule
\end{tabularx}
\\[1.0ex](D)
\end{minipage}
\vspace{-4mm}
\label{tab:one_table}
\end{table*}

\noindent\textbf{Number of Initial Key-Frames.}
We investigate the impact of the starting point of our progressive refinement by varying the number of initial key-frames as shown in Table~\ref{tab:start_num_frame}.
This result highlights a central principle of our approach that it is not the total amount of input frames that matters but rather the ability to insert new frames only when they are truly needed.
Starting with a minimal number of frames (i.e., two) and selectively adding more based on gradient signals enables the model to focus on semantically meaningful motion without being distracted by redundant or misaligned updates.
The performance drop observed when increasing the number of initial frames supports this idea by showing that na\"ively including more content can actually interfere with the optimization process.
We hypothesize that starting with too many frames creates conflicting gradient signals in the early training stages, which hinders the optimization of a coherent motion trajectory.
In contrast, our ``start-small" approach allows the model to first establish a robust and simple path between two anchors before refining more complex, non-linear dynamics.

\noindent\textbf{Without Frame Insertion.}
Removing the adaptive insertion leads to a substantial performance drop as shown in Table~\ref{tab:one_table} (A).
This result confirms that static interpolation between key-frames fails to capture the non-linear temporal dynamics inherent in complex actions.
The necessity of this component aligns with our theoretical framework where additional key-frames are required to resolve gradient misalignment at intermediate timestamps.
Even in this restricted state, the holistic baseline remains competitive against prior work, which underscores the strength of our underlying interpolation architecture.

\noindent\textbf{Frame Selection Strategy.}
We evaluate the effectiveness of our gradient-guided frame selection by comparing it against a random baseline under identical insertion triggers.
As detailed in Table~\ref{tab:one_table} (B), replacing our targeted selection with random addition results in inferior performance across both datasets.
This indicates that the specific directional misalignment of gradients effectively points to temporal junctions that require additional representational capacity.
Gradient correlation serves as a decisive signal for identifying semantically meaningful frames rather than just increasing the total frame count.

\noindent\textbf{Cosine Similarity vs. L2 Distance.}
Directional misalignment is a more reliable indicator of frame-level variation than simple distance-based metrics.
We validate this by comparing cosine similarity against an L2 distance alternative calibrated to a comparable geometric threshold.
The superior results of the angle-based criterion in Table~\ref{tab:one_table} (C) suggest that the relative orientation of gradient vectors better captures the semantic transitions necessary for frame insertion.
This justifies our focus on directional disagreement as the primary trigger for expanding the synthetic video set.

\noindent\textbf{Effect of Warm-Up and Cool-Down Phases.}
Stable condensation dynamics require a structured training curriculum consisting of both warm-up and cool-down phases.
Removing the warm-up phase introduces early gradient noise that destabilizes the initial optimization of the motion trajectory.
Conversely, the absence of a cool-down phase leads to over-insertion and noisy representations as the model nears convergence.
The results in Table~\ref{tab:one_table} (D) confirm that both stages are integral to ensuring the model remains focused on a coherent and efficient representation.

\noindent\textbf{Action Retrieval Task.}
\begin{table}[t!]
\caption{Action retrieval performance (Recall@K) on the HMDB51 dataset under the 1 VPC setting.}
\vspace{-2mm}
\centering
\small
\begin{tabularx}{\columnwidth}{YYYYYYYY}
\toprule
\multicolumn{4}{c}{Wang et al.} & \multicolumn{4}{c}{PRISM} \\
\cmidrule(lr){0-3} \cmidrule(lr){5-8}
R@1 & R@5 & R@10 & R@20 & R@1 & R@5 & R@10 & R@20 \\
\cmidrule(lr){0-3} \cmidrule(lr){5-8}
22.61 & 40.92 & 52.22 & 65.82 & 37.97 & 56.14 & 66.14 & 77.65 \\
\bottomrule
\end{tabularx}
\vspace{-4mm}
\label{tab:action_retrieval}
\end{table}
Effective dataset condensation requires synthesized data to capture intrinsic semantic properties that generalize beyond action recognition task.
We validate the practicality of PRISM by evaluating the condensed representations on an action retrieval task.
This involves training an embedding model with a metric learning objective using the 1 VPC HMDB51 dataset.
As shown in Table~\ref{tab:action_retrieval}. the substantial margin over the prior distillation baseline demonstrates that PRISM is not strictly bounded to a specific task.
This transferability suggests that the adaptively inserted frames capture essential spatiotemporal cues providing a plausible candidate for leveraging PRISM in applications beyond action recognition task.
\section{Conclusion}
\label{sec:conclusion}

We introduced Progressive Refinement and Insertion for Sparse Motion (PRISM), a paradigm for video dataset condensation that preserves the critical interplay between content and motion.
Our method dynamically synthesizes key-frames where they are most needed by identifying gradient misalignments that signal high motion complexity.
This adaptive strategy allows the model to allocate representational capacity intelligently, resulting in compact and temporally coherent condensed datasets.
Extensive experiments demonstrate that PRISM achieves competitive performance and state-of-the-art storage efficiency across standard action recognition benchmarks.

\noindent\textbf{Limitations.}
The performance of PRISM may be constrained in videos with extremely abrupt motion where linear interpolation between key-frames is insufficient.
Optimization from noise can also become unstable for very long sequences exceeding 16 frames.
Although we have demonstrated the effectiveness of PRISM in both action recognition and action retrieval tasks, its ability to preserve the fine-grained semantic cues required for complex generative tasks remains unexplored.
Future research could incorporate quantitative motion fidelity metrics to provide a more rigorous evaluation of the synthesized dynamics.

\newpage

\noindent\textbf{Acknowledgements}
This work was supported by Institute of Information \& communications Technology Planning \& Evaluation (IITP) grant funded by the Korea government(MSIT) (No. RS-2022-II0951, Development of Uncertainty-Aware Agents Learning by Asking Questions)
and
Institute of Information \& Communication Technology Planning \& Evaluation (IITP) grant funded by the Korea government (MSIT) (No. RS-2025-02263841, Development of a Real-time Multimodal Framework for Comprehensive Deepfake Detection Incorporating Common Sense Error Analysis)
and
Artificial intelligence industrial convergence cluster development project funded by the Ministry of Science and ICT(MSIT, Korea)\&Gwangju Metropolitan City.
{
    \small
    \bibliographystyle{ieeenat_fullname}
    \bibliography{main}
}
 
\clearpage
\setcounter{page}{1}
\maketitlesupplementary
\setcounter{section}{0}
\setcounter{subsection}{0}
\setcounter{table}{0}
\setcounter{figure}{0}
\setcounter{lemma}{0}

\renewcommand{\thesection}{\Alph{section}}
\renewcommand{\thesubsection}{\Alph{subsection}}
\renewcommand{\thetable}{\Alph{table}}
\renewcommand{\thefigure}{\Alph{figure}}

\subsection{Proof of Lemma~\ref{lem:loss_blockage}}
\label{supple:proof}

\begin{lemma}\textnormal{\textbf{(Loss-Descent Blockage under Gradient Misalignment)}}

\noindent\label{lem:loss_blockage}
Let \(s_t = \alpha s_{k_i} + (1 - \alpha) s_{k_{i+1}}\), with \(0 < \alpha < 1\), be a linearly interpolated frame between two key-frames \(s_{k_i}\) and \(s_{k_{i+1}}\).
Let the task-loss gradients be denoted as
\[
  g_t = \nabla_{s_t} \mathcal{L}(s_t), \quad
  g_i = \nabla_{s_{k_i}} \mathcal{L}(s_{k_i}),
\]
\[
  g_{i+1} = \nabla_{s_{k_{i+1}}} \mathcal{L}(s_{k_{i+1}}).
\]
Suppose
\[
  \langle g_t,\, g_i \rangle < 0
  \quad \text{and} \quad
  \langle g_t,\, g_{i+1} \rangle < 0.
\]
Then, for every convex combination
\[
  v = \lambda (-g_i) + (1 - \lambda)(-g_{i+1}), \quad \lambda \in [0, 1],
\]
the following inequality holds:
\[
  \langle g_t,\, v \rangle > 0.
\]
Consequently, no first-order update obtained by modifying only the two endpoint frames can decrease \(\mathcal{L}\) at \(s_t\); the loss is stationary or strictly increasing along every such direction. Therefore, \(s_t\) must be promoted to the key-frame set and directly optimized to enable further loss minimization.
\end{lemma}

\emph{Proof.} By the bilinearity of the inner product,

\[
    \langle g_t, v \rangle = \lambda \langle g_t, -g_i \rangle + (1 - \lambda) \langle g_t, -g_{i+1} \rangle.
\]

Applying the assumption \(\langle g_t, g_i \rangle < 0\), we obtain

\[
\langle g_t, -g_i \rangle = -\langle g_t, g_i \rangle > 0,
\]

and similarly,

\[
\langle g_t, -g_{i+1} \rangle = -\langle g_t, g_{i+1} \rangle > 0.
\]

Therefore,

\[
\langle g_t, v \rangle
= \lambda \cdot \langle g_t, -g_i \rangle + (1 - \lambda) \cdot \langle g_t, -g_{i+1} \rangle
> 0.
\]

This shows that the directional derivative of \(\mathcal{L}\) at \(s_t\) along any direction \(v\) formed by adjusting only the endpoints is positive:

\[
D_{v} \mathcal{L}(s_t) = \langle \nabla \mathcal{L}(s_t), v \rangle > 0.
\]

Thus, no first-order update along such directions can reduce the loss at \(s_t\), and \(\mathcal{L}(s)\) is strictly increasing along all directions spanned by \(-g_i\) and \(-g_{i+1}\). It follows that further loss minimization requires directly optimizing \(s_t\) as a key-frame. \hfill \(\square\)

\subsection{Hyperparameter}
\label{supple:hyperparameter}
\begin{table*}[h]
\centering
\small
\caption{Hyperparameters for PRISM under different datasets and IPC.}
\label{tab:supple_hyperparameter}
\begin{tabularx}{\textwidth}{llYYYYY}
\toprule
\multirow{2}{*}{Method} & \multirow{2}{*}{Dataset} & \multicolumn{3}{c}{Train} & \multicolumn{2}{c}{Evaluation}  \\
\cmidrule(lr){3-5} \cmidrule(lr){6-7}
& & IPC & LR & Batch Real & Epoch & LR \\
\midrule
\multirow{8}{*}{PRISM} & \multirow{3}{*}{MiniUCF} & 1 & 1 & 64 & \multirow{10}{*}{500} & \multirow{10}{*}{$1e^{-2}$} \\
& & 5 & 25 & 64 & & \\
& & 10 & 50 & 64 & & \\
[0.1ex] \cdashline{2-5}[3pt/3pt] \\ [-1.8ex]
& \multirow{3}{*}{HMDB51} & 1 & 0.7 & 64 & & \\
& & 5 & 25 & 64 & & \\
& & 10 & 75 & 64 & & \\
[0.1ex] \cdashline{2-5}[3pt/3pt] \\ [-1.8ex]
& \multirow{2}{*}{Kinetics-400} & 1 & 1 & 64 &  & \\
& & 5 & 50 & 128 & & \\
[0.1ex] \cdashline{2-5}[3pt/3pt] \\ [-1.8ex]
& \multirow{2}{*}{SSv2} & 1 & 3 & 64 & & \\
& & 5 & 30 & 128 & & \\
\bottomrule
\end{tabularx}
\end{table*}

In Table~\ref {tab:supple_hyperparameter}, we show the learning rate and batch size under each dataset and IPC.
The \(\epsilon\) is set to 0 for all experiments throughout the manuscript.
The warm-up and cool-down phases are processed for 20\% of the whole iteration each.
In other words, if the condensation process is set to 100 iterations, the warm-up phase takes up the first 20 iterations and the cool-down phase takes up the last 20 iterations, leaving 80 iterations for the progressive refinement and insertion of frames.
We follow the setting from the prior method~\citep{video_distillation_2024} for evaluation and cross-architecture evaluation.

\paragraph{Justification for experimental scale.} We emphasize that our experimental setup, using a lightweight backbone, miniC3D, low resolutions (e.g., 64x64), and a small number of frames (T=8 or T=16), follows the established protocol from prior work in video dataset condensation~\cite{video_distillation_2024}. This constrained setting is a necessary consequence of the task's extreme computational demands.
Unlike standard model training, dataset condensation requires an iterative optimization process to synthesize the data itself.
This process, whether through gradient matching or distribution matching, requires repeated forward and backward passes through the network to update the synthetic data, making it orders of magnitude more costly than a standard training epoch.
Therefore, using larger models (e.g., I3D), high resolutions (e.g., 224x224), or long sequences is computationally prohibitive for current condensation methods.
Our goal is thus to demonstrate the relative efficacy and efficiency of the condensation method within this feasible, standardized environment.

\paragraph{Effect of Initialization Strategy}
We validate our ``start-small" initialization approach by comparing it against a variant where a full-length video is interpolated from Gaussian noise endpoints, and all interpolated frames are treated as learnable parameters from the very beginning.
Experimental results show that this fully-learnable initialization severely degrades performance, yielding only 5.69\% on HMDB51 and 15.15\% on miniUCF at 1 VPC.
We observe that optimization becomes temporally imbalanced in this setting; gradient updates concentrate heavily on the middle frames, leaving the endpoint frames weakly optimized. This supports our design choice to start with a minimal anchor set and allocate learnable capacity progressively only when gradient conflicts indicate true non-linear motion.

\begin{figure*}[t!]
  \centering
   \includegraphics[width=1.95\columnwidth]{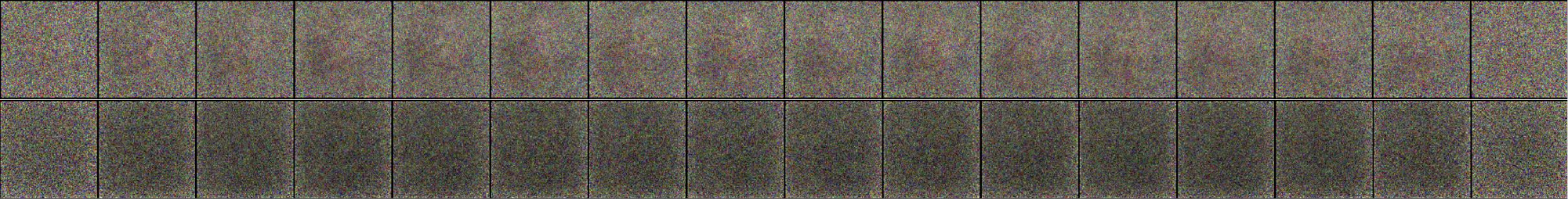}
   \caption{Visualized sequences of the condensed frames.}
   \label{fig:all_learnable}
\end{figure*}

\subsection{$\epsilon$ Sweep}
\begin{table*}[t!]

\caption{$\epsilon$ Sweep on MiniUCF and HMDB51 dataset on 1 IPC.}

\begin{tabularx}{\textwidth}{YYYYYYYY}
\toprule
\multirow{2}{*}{Dataset} & \multicolumn{7}{c}{$\epsilon$} \\
\cmidrule{2-8}
& -0.3 & -0.2 & -0.1 & 0 & 0.1 & 0.2 & 0.3 \\
\midrule
MiniUCF & 16.2 & 16.6 & 16.8 & \textbf{17.9} & 17.6 & 17.4  & 16.9\\
HMDB51  & 6.4 & 6.5 & 6.8 & \textbf{7.5} & 7.4 & 6.8 & 6.6 \\
\bottomrule
\end{tabularx}
\label{tab:epsilon_sweep}
\end{table*}
The sweep over $\epsilon$ reveals a clear unimodal trend centered at $\mathbf{\epsilon=0}$ which produces the highest accuracy across both miniUCF and HMDB51 as shown in Table ~\ref{tab:epsilon_sweep}. This value is critical because it represents the point where the cosine similarity transitions from a positive to a negative correlation.
When we apply the insertion criterion $\cos_i^t < \epsilon$ with $\mathbf{\epsilon=0}$, the condition for insertion becomes $\mathbf{\cos(\cdot) < 0}$.
Geometrically, this requires the gradient $\mathbf{g_t}$ of the interpolated frame to be pointing into the opposite half-space (i.e., making an angle greater than $90^\circ$) relative to both adjacent key-frame gradients $\mathbf{g_i}$ and $\mathbf{g_{i+1}}$.
This directional disagreement is the most direct signal that the interpolated frame $s_t$ lies in a region of the loss landscape where the current linear interpolation between the key-frames $\mathbf{(s_{k_i}, s_{k_{i+1}})}$ cannot model the motion. The existence of an interpolated point whose optimal update direction is pointing away from the update directions of its anchors signifies a sharp, non-linear change in the spatiotemporal content, demanding the insertion of a new key-frame.

If $\epsilon$ is too negative (e.g., $-0.3$), the criterion becomes overly strict, allowing only the most severely misaligned gradients to trigger insertion.
As a result, too few key-frames are added, limiting PRISM's ability to capture non-linear motion transitions, a phenomenon consistent with the ``without insertion'' ablation (Table 6(A) in the main manuscript) that demonstrated reduced performance.
Conversely, when $\epsilon$ is positive (e.g., $0.2$ or $0.3$), the threshold becomes overly permissive, causing many interpolated frames with mildly differing gradients to be unnecessarily promoted.
This leads to noisy or redundant key-frame sets and destabilizes optimization.
Thus, $\epsilon=0$ provides the ideal, non-arbitrary balance by defining the precise point of directional disagreement that signals non-linearity.

\subsection{Warm-Up and Cool-Down Sensitivity Sweep}
\begin{table*}[t!]
\caption{Warm-Up and Cool-Down Sensitivity on MiniUCF and HMDB51 dataset on 1 IPC.}
\begin{tabularx}{\textwidth}{YYYYY}
\toprule
Dataset &  0.3 & 0.2 & 0.1 & 0\\
\midrule
MiniUCF & 17.6 & \textbf{17.9} & 17.4 & 17.3 \\
HMDB51  & 7.3 & \textbf{7.5} & 7.1 & 6.8 \\
\bottomrule
\end{tabularx}
\label{tab:wr_sweep}
\end{table*}
The warm-up and cool-down sensitivity sweep shows that intermediate scheduling (0.2) yields the strongest performance, while both extremes degrade results.
This pattern reflects PRISM’s reliance on stable insertion dynamics throughout training.
A longer warm-up (i.e., 0.3) delays frame insertion excessively, preventing PRISM from capturing early non-linear transitions and resulting in under-refinement of the condensed video.
On the other hand, removing warm-up entirely (i.e., 0) allows frames to be inserted before the initial key-frames have stabilized exposing insertion decisions to noisy early gradients that lead to suboptimal or redundant frames being selected.
Similarly, an extended cool-down prematurely stops insertion and prevents late, meaningful refinements, while removing cool-down entirely allows late insertions that do not receive enough optimization steps, degrading the final representation.
The peak at 0.2 arises because it provides a balanced curriculum—early enough stabilization to avoid noisy insertions and late enough freezing to prevent under-optimized or overly large key-frame sets.
This balance mirrors the observed necessity of both warm-up and cool-down in the main ablation, further validating PRISM’s temporal curriculum.

\subsection{Qualitative Results}
\label{supple:qualitative}

\begin{figure*}[t]
  \centering
   \includegraphics[width=\textwidth]{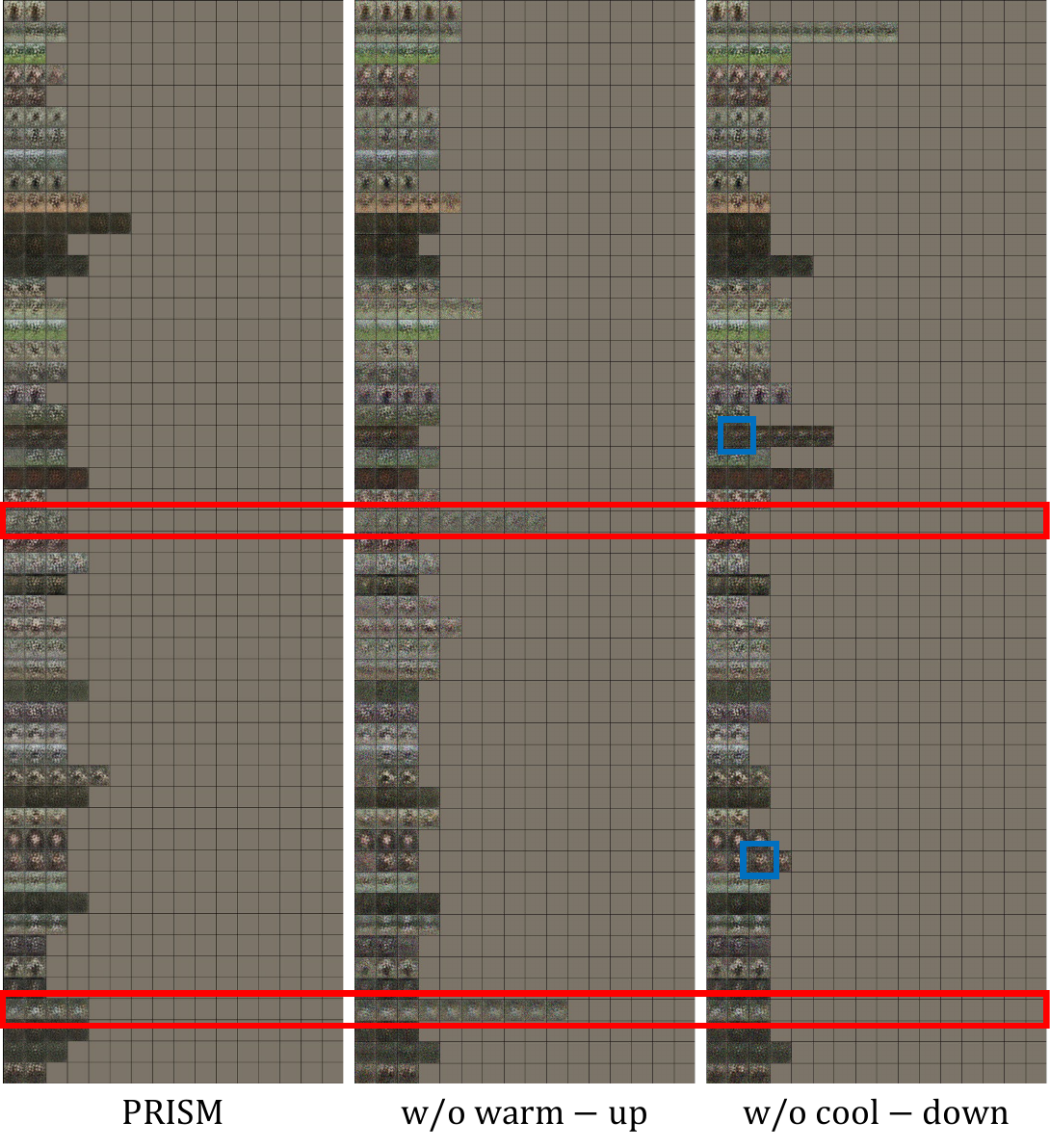}
   \caption{Visualization of PRISM, PRISM without warm-up, and PRISM without cool-down on HMDB51 under 1 VPC. Red rectangles highlight the negative effects of omitting the warm-up phase, while blue rectangles indicate frames that may be under-trained due to the absence of a cool-down phase.}
   \label{fig:supple_1}
\end{figure*}

\begin{figure*}[t]
  \centering
   \includegraphics[width=\textwidth]{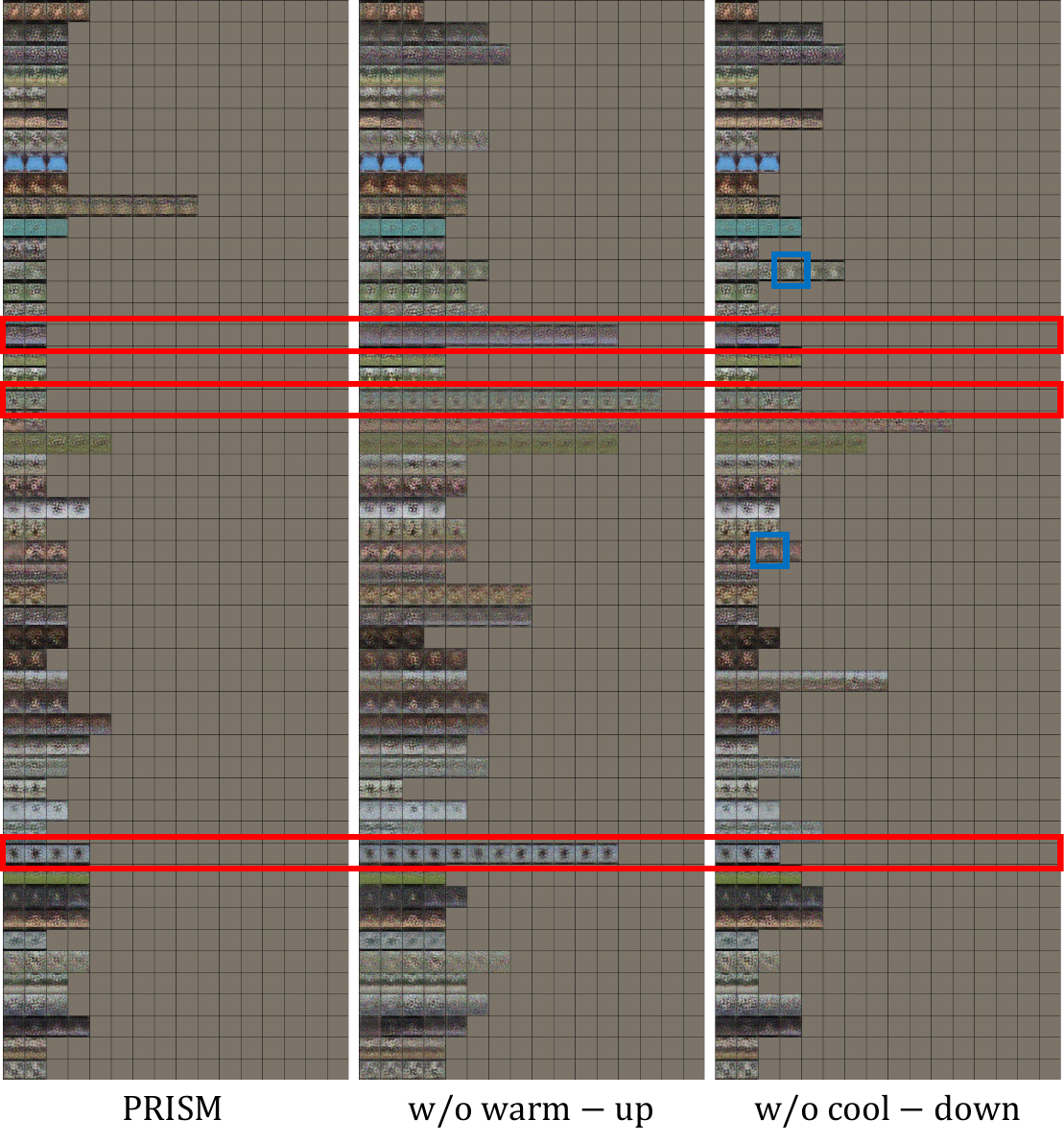}
   \caption{Visualization of PRISM, PRISM without warm-up, and PRISM without cool-down on MiniUCF under 1 VPC. Red rectangles highlight the negative effects of omitting the warm-up phase, while blue rectangles indicate frames that may be under-trained due to the absence of a cool-down phase.}
   \label{fig:supple_2}
\end{figure*}

We visualize the condensed videos on HMDB51 and MiniUCF under the 1 VPC setting for maximal clarity.
The visualized frames in Figure~\ref{fig:supple_1} and Figure~\ref{fig:supple_2} correspond to those retained after the condensation process, where the noise images are placeholders which does not get stored along with condensed data.

Red rectangles highlight the negative effect when the warm-up phase is omitted.
As consistently observed across both datasets, removing the warm-up leads to excessive frame selection, resulting in redundant and less informative synthetic frames while consuming more memory.

Blue rectangles indicate frames produced when the cool-down phase is omitted.
Although overall results appear more stable than in the warm-up-removed case, we observe that some frames are added during the final few iterations of condensation.
These late-added frames often lack sufficient training, reducing their utility for action recognition by being not fully trained.

\subsection{The Use of Large Language Models (LLMs)}

We used a Large Language Model (LLM) to assist with improving the clarity, grammar, and organization of the text. All scientific contributions, including the core methodology, experimental design, and analysis of results, are solely the work of the authors.

\end{document}